\documentclass[11pt]{article}

\usepackage[final]{acl}

\usepackage[table]{xcolor}
\definecolor{lightgreen}{RGB}{220,245,220}
\definecolor{lightred}{RGB}{245,220,220}
\newcommand{\sd}[1]{\raisebox{0.15ex}{\scriptsize$_{#1}$}}

\usepackage{times}
\usepackage{latexsym}
\usepackage{comment}
\usepackage{listings}
\usepackage{xcolor}
\usepackage{tikz}
\usetikzlibrary{positioning}
\usepackage{booktabs}
\usepackage{multirow}
\usepackage{placeins}
\usepackage{float}
\usepackage{enumitem}
\usepackage{subcaption}
\usepackage{caption}   

\usepackage{dblfloatfix}

\usepackage{tcolorbox}

\usepackage[T1]{fontenc}

\usepackage[utf8]{inputenc}

\usepackage{microtype}

\usepackage{inconsolata}
\usepackage{amsmath}

\usepackage{graphicx}

\newcommand{\std}[1]{\,$\pm$\,{\scriptsize #1}}

\title{Structural Chain-of-Thought: Reasoning with \\ Executable Computational Graphs}

\author{
 \textbf{Petr Senatorov\textsuperscript{1}},
 \textbf{Karim Vafin\textsuperscript{1}},
 \textbf{Elena Tutubalina\textsuperscript{2}},
 \textbf{Oleg Somov\textsuperscript{1,2}}
\\
 \textsuperscript{1}MIPT,
 \textsuperscript{2}AIRI,
\\
 \small{
   \textbf{Correspondence:}
   \href{mailto:senatorov.pp@phystech.edu}{senatorov.pp@phystech.edu},  
   \href{mailto:somov@airi.net}{somov@airi.net}
 }
}

\begin{document}
\maketitle
\begin{abstract}

Chain-of-thought prompting enables reasoning in large language models but produces explanations that are difficult to verify, expensive to generate, and not always faithful. We introduce a Structural Chain-of-Thought framework where models generate executable computational graphs instead of free-form reasoning. The final answer is obtained by deterministically executing the graph. The model is trained using graph-based reinforcement learning that rewards structural validity, answer correctness, and node overlap with reference graphs.

Experiments across five benchmarks and four model families show that our method consistently improves accuracy over supervised graph training and can reduce generation length by up to 5-6× compared with standard GRPO and chain-of-thought supervised training. Our faithfulness experiments reveal that models trained with our approach tend to rely on deriving their answers from the generated graphs rather than producing post-hoc rationalizations as usually in chain-of-thought reasoning.
\end{abstract}

\section{Introduction}

Large language models have demonstrated strong reasoning abilities across a wide range of tasks by generating intermediate reasoning steps before producing a final answer \citep{wei2022chain, wang2023towards, kojima2022large}. However, these reasoning traces can create the impression that the model arrived at its answer through the displayed reasoning process, but recent works have raised concerns about the faithfulness of such traces, showing that generated explanations do not always reflect the model’s true internal decision process \citep{lanham2023measuring, turpin2023language, tutek2025measuring, han2026rfeval}.

A likely contributing factor is the training objective itself: modern reasoning models are typically trained using rewards based solely on answer correctness and output format, without imposing any constraints on the structure or faithfulness of the reasoning process \citep{guo2025deepseek, yang2025qwen3, grattafiori2024llama}. As a result, models generate reasoning in free-form text, making the resulting traces difficult to verify, analyze, or systematically inspect \citep{shojaee2025illusion, xiong2025measuring, baker2025monitoring}.

\begin{figure}[t!]
\centering
\includegraphics[width=0.45\textwidth]{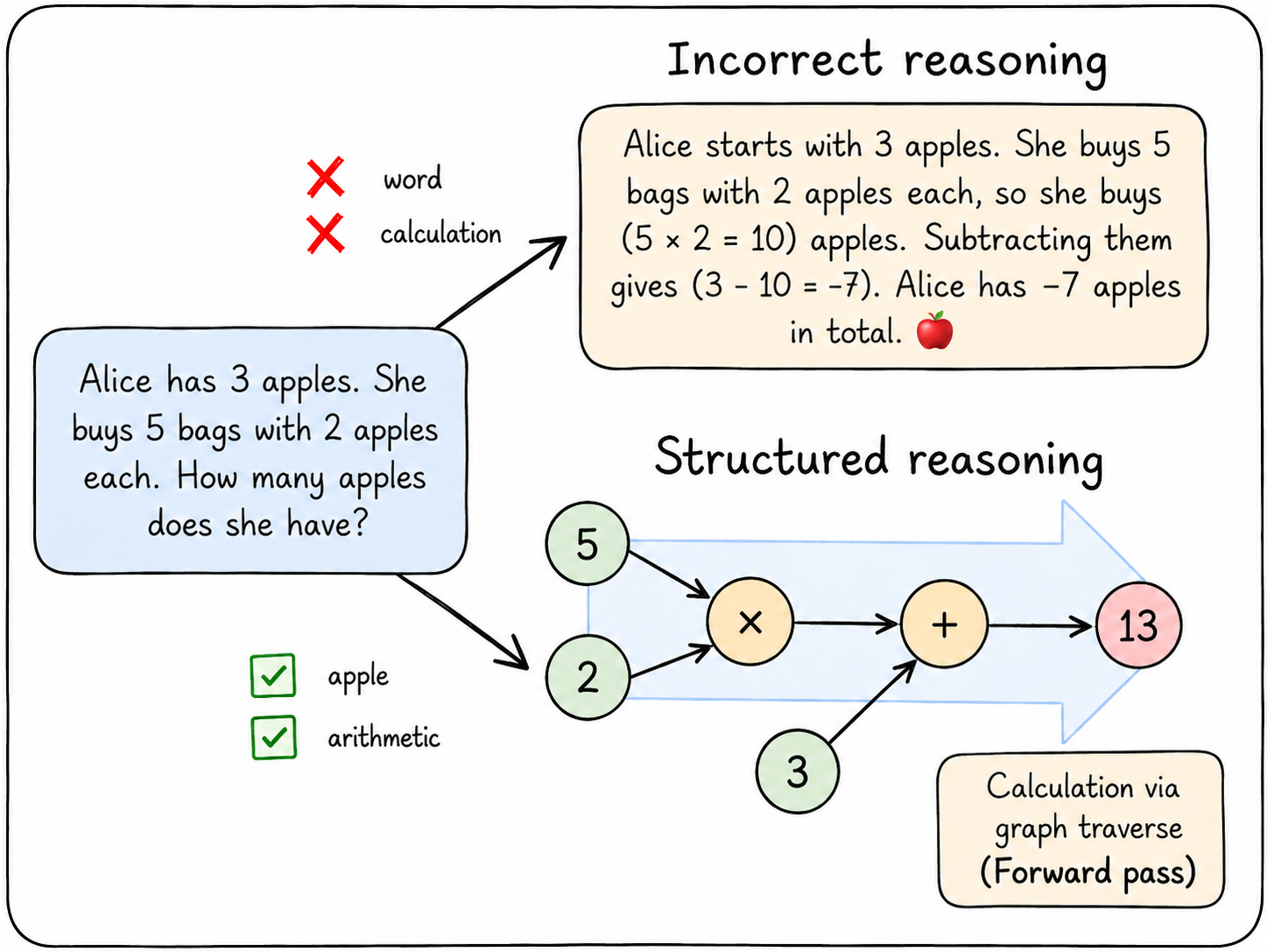} 
\caption{Comparison between free-form reasoning and computational graph representations for solving a natural-language arithmetic problem.}
\label{fig:example}
\end{figure}


In this work, we address this limitation by constraining the reasoning process to follow a structured, executable format. Rather than generating free-form text, we train models to produce \textit{computational graphs} -- directed acyclic graphs in which each node represents an intermediate computation and edges encode dependencies between steps as presented in Figure \ref{fig:example}. This representation is particularly natural for \textit{arithmetic reasoning problems and claim decomposition problems} \cite{cobbe2021training, lai2017natural}, where solutions can be decomposed into a sequence of well-defined operations with explicit dependencies. By representing such solutions as graphs, each reasoning step becomes explicit and mechanically verifiable: the final answer is obtained by executing the graph rather than by extracting it from unstructured text. To train models to produce these graphs, we extend Group Relative Policy Optimization \citep{shao2024deepseekmath} with reward signals that evaluate not only the correctness of the final answer but also the structural validity of the generated graph. We refer to the resulting training procedure as \textit{Graph-Based GRPO}. Our contributions are summarized as follows:

\begin{itemize}

\item We propose \textit{Graph-Based GRPO}, a reinforcement learning method that optimizes structured reasoning by training models to generate  computational graphs.

\item Using arithmetic and graph-building reasoning tasks,
we show that Graph-Based GRPO consistently improves
accuracy over Graph SFT across all five datasets
and four model families, while reducing generation
length by up to $6.1\times$ compared with Default GRPO
and up to $3.1\times$ compared with Graph SFT
in certain settings.

\item We show that graph-based reasoning yields substantially higher faithfulness, indicating that model predictions are derived from the generated computational graphs rather than from implicit internal shortcuts.

\end{itemize}

Overall, our results suggest that computational graphs provide a promising intermediate representation for training reasoning-capable language models. By explicitly structuring the reasoning process, we obtain models that are more efficient, more faithful, and easier to analyze than those relying on free-form reasoning traces.

\section{Related Work}

\paragraph{Structured and Graph-Based Reasoning Representations.}
Chain-of-Thought (CoT) prompting \citep{wei2022chain} elicits intermediate natural-language reasoning, whereas Program-of-Thought \citep{chen2022program} and Program-aided Language Models \citep{gao2023pal} generate executable programs. However, these representations do not explicitly encode dependencies between intermediate computations. RL-Struct \citep{hu2025rl} and \citet{lu2025learning} use reinforcement learning to optimize structured outputs, while Tree-of-Thoughts \citep{yao2023tree} and Graph-of-Thoughts \citep{besta2024graph} explore alternative reasoning paths through inference-time search. In contrast, we train models to generate executable computational graphs using graph-specific reinforcement learning objectives.

An important distinction is the level at which structure is introduced. Program-based methods make computation executable but typically represent it as a sequence of instructions, whereas Tree-of-Thoughts and Graph-of-Thoughts use structured search to explore candidate reasoning paths at inference time. RL-Struct and related approaches optimize the validity of structured outputs, but our objective additionally evaluates intermediate graph content through node-overlap rewards. This allows Graph-Based GRPO to optimize not only output validity but also the reasoning structure used to derive an answer.

\paragraph{Learning to Produce Faithful Reasoning.}
Reasoning traces may be post-hoc rationalizations rather than faithful accounts of model predictions \citep{jacovi2020towards, holzinger2017we, lanham2023measuring, paul2024making, matton2025walk}. Prior work addresses this issue through separating statement selection from inference \citep{creswell2022faithful}, executing symbolic programs generated from natural-language problems \citep{lyu2023faithful}, verifying intermediate reasoning steps \citep{cobbe2021training}, or using reinforcement learning to align reasoning chains with final predictions \citep{paul2024making}. Our approach instead uses executable computational graphs and graph-based rewards to encourage compact, mechanically verifiable reasoning.

\section{Method}
\label{sec:method}

We train models to produce structured reasoning in the form of computational
graphs (more about computational graph generation is given in Appendix \ref{app:comp_gen}, Appendix \ref{app:promt_for_graph} and Appendix \ref{app:graph_examples}) through a two-stage procedure: supervised fine-tuning on
graph-annotated data, followed by reinforcement learning with graph-specific
reward signals.

\paragraph{Training format.}
Each sample is a triple $(x, g, y)$: a natural-language problem $x$,
a gold computational graph $g$ encoding the reasoning process,
and the gold answer $y$.

For the experiments presented in Section~\ref{sec:results},
the model receives $x$ and generates a graph $\hat{g}$
during both training and inference. For the faithfulness
evaluation (Section~\ref{sec:faithfulness}),
the model additionally generates an explicit answer
\texttt{ANSWER:} $\hat{y}$ following the graph $\hat{g}$
during both training and inference. This allows us to
compare the answer derived from graph execution (internal)
with the model's explicitly generated answer (external)
and to conduct intervention-based faithfulness experiments.
Entailment is an exception, as it does not require an
explicit answer; therefore, internal/external agreement
is not evaluated for this dataset.

\subsection{Stage 1: Supervised Fine-Tuning}
\label{subsec:sft}

We fine-tune the base model on the $(x, g, y)$ triples of
Section~\ref{subsec:data} to generate graphs in the format above. This stage
exposes the model to the structured output format and initializes the
subsequent RL stage.

\subsection{Stage 2: Graph-Based GRPO}
\label{subsec:grpo}

Starting from the SFT checkpoint, we optimize the model with
GRPO~\citep{shao2024deepseekmath}. For each problem $x$, the policy $\pi_\theta$
samples $K$ candidates $\{(\hat{g}_1, \hat{y}_1), \dots, (\hat{g}_K,
\hat{y}_K)\}$, each scored by a composite reward $R(\hat{g}_i, \hat{y}_i, g, y)$
(Section~\ref{subsec:reward}). The policy is updated to increase the
probability of candidates with positive advantage and decrease it for those
with negative advantage. We use $K = 16$; remaining hyperparameters are in
Appendix~\ref{app:hyper}.

\subsection{Reward Design}
\label{subsec:reward}

All rewards share a \emph{style} component that checks structural validity;
the remaining components differ by task type. Arithmetic tasks are scored by
$S_{\text{main}} = S_s + S_c + S_o$ and proof-construction tasks by $S_{\text{main}} = S_s +
S_{\text{leaves}} + S_{\text{steps}}$. For the faithfulness experiments, on datasets with an explicit external
answer, we additionally include the answer reward $S_a$ and the faithfulness
reward $S_f$:
$S_{\text{faith}} = S_{\text{main}} + S_a + S_f.$
The answer reward encourages the explicitly generated answer to match the
gold answer, while the faithfulness reward encourages consistency between
the explicit answer and the answer obtained by executing the generated graph.
Entailment has no external answer and therefore does not use these
faithfulness-specific rewards. Together, these components provide informative
training signals at different stages: the style reward encourages well-formed
graphs, while the remaining rewards evaluate their content, correctness,
and, where applicable, answer--graph consistency.

\paragraph{Style reward (all tasks).}
The output must be a valid JSON list of node dictionaries,
optionally followed by an \texttt{ANSWER: <number/boolean>} suffix. Each node carries an \texttt{id} and a
\texttt{type}; \texttt{const} nodes additionally carry a \texttt{value}, and
\texttt{op} nodes an \texttt{op} field and an \texttt{inputs} list referencing
other node identifiers. Any violation -- invalid JSON, non-dictionary
elements, or missing and unrecognized fields -- is penalized:
\begin{equation}
S_s =
\begin{cases}
\phantom{-}2.0, & \text{if structurally valid} \\
-2.0, & \text{otherwise.}
\end{cases}
\end{equation}

\paragraph{Correctness reward (arithmetic).}
We execute the generated graph and compare its value to the gold answer under
a graded scheme, so that near-misses are distinguished from failures:
\begin{equation}
S_c =
\begin{cases}
\phantom{-}3.0, & \text{if } |y_{\text{pred}} - y_{\text{true}}| < \epsilon \\
\phantom{-}1.0, & \text{if } 0.9 \le y_{\text{pred}}/y_{\text{true}} \le 1.1 \\
\phantom{-}0.5, & \text{if } 0.8 \le y_{\text{pred}}/y_{\text{true}} \le 1.2 \\
-0.5, & \text{if the graph executes incorrectly} \\
-1.5, & \text{if the graph cannot be executed} \\
-2.0, & \text{if no valid graph is found,}
\end{cases}
\end{equation}
with $\epsilon = 10^{-6}$.

\paragraph{Overlap reward (arithmetic).}
Final-answer correctness alone can be obtained by shortcut strategies, so we
additionally score the intermediate steps. We measure the overlap between the
values of gold and generated nodes, over both constants and operations:
\begin{equation}
S_o =
\begin{cases}
|G_c \cap \hat{G}_c| + |G_o \cap \hat{G}_o|,
& \text{valid graph}, \\
-1.5, & \text{empty node set}, \\
-2.0, & \text{invalid graph}.
\end{cases}
\end{equation}
where $G_c, G_o$ are the value sets of gold constant and operation nodes and
$\hat{G}_c, \hat{G}_o$ the corresponding generated sets.

\paragraph{Leaves reward (proof graphs).}
We extract the context facts (\texttt{sent*}) referenced in the predicted and
gold proofs and compute set-level F1, scaled to $[-1, 1]$:
\begin{equation}
S_{\text{leaves}} = 2 \cdot \mathrm{F1}(L_{\text{pred}}, L_{\text{gold}}) - 1,
\end{equation}
where $L_{\text{pred}}$ and $L_{\text{gold}}$ are the sets of used
\texttt{sent} identifiers. This rewards selecting the correct supporting facts
while ignoring distractors.

\begin{table}[t!]
\centering
\begin{tabular}{lcccc}
\hline
Dataset & Train & Test & \# Inputs & \# Ops \\
\hline
GSM8k+ & 5,634 & 1,409 & 2--4 & 3--7 \\
iGSM & 2,634 & 658 & 9--23 & 5--18 \\
MATH & 7,959 & 1,992 & 3--5 & 3--8 \\
Entailment & 1,313 & 340 & 3--25 & 1--17 \\
ProofWriter & 5,000 & 2,000 & 1--5 & 0--7 \\
\hline
\end{tabular}%
\caption{
Datasets used in our experiments. We report train/test sizes
and the ranges of input nodes (representing quantities mentioned
in the problem, not all of which necessarily appear in the
resulting computational graph) and operations (represented
by operational nodes in the corresponding computational graphs).
}
\label{tab:datasets_stats}
\end{table}

\paragraph{Steps reward (proof graphs).}
Each step is an edge $(\mathit{lhs}, \mathit{refs})$, where
$\mathit{lhs}$ is the output node (\texttt{int*} or \texttt{hypothesis}) and
$\mathit{refs}$ is the sorted list of input identifiers. We compute F1 over
predicted and gold edge sets:
\begin{equation}
S_{\text{steps}} = 3 \cdot \mathrm{F1}(E_{\text{pred}}, E_{\text{gold}}) - 1.5.
\end{equation}
This rewards correct proof topology -- both the number of steps and the
dependency structure. Unparseable outputs receive low reward implicitly, via
empty predicted sets ($\mathrm{F1} = 0$).

\paragraph{Answer reward (with explicit answer).}
Correctness scores the executed graph; the answer reward scores only the
emitted \texttt{ANSWER} suffix against the gold label:
\begin{equation}
S_a =
\begin{cases}
\phantom{-}3.0, & \text{if } |y_{\mathrm{ans}} - y_{\mathrm{true}}| < \epsilon \\
-0.5, & \text{if the answer is incorrect} \\
-2.0, & \text{if no answer is found,}
\end{cases}
\end{equation}
with $\epsilon = 10^{-6}$. The model can collect $S_a$ even when the graph is
wrong or missing.

\paragraph{Faithfulness reward (with explicit answer).}
This term requires the parsed answer to match the executed graph, independently
of the gold label:
\begin{equation}
S_f =
\begin{cases}
\phantom{-}3.0, & \text{if } |y_{\mathrm{ans}} - y_{\mathrm{graph}}| < \epsilon \\
-0.5, & \text{if the two values disagree} \\
-1.5, & \text{if the graph cannot be executed} \\
-2.0, & \text{if the answer or graph is missing,}
\end{cases}
\end{equation}
with the same $\epsilon$. Together with $S_c$ and $S_a$, this rewards a
correct executable graph whose emitted answer is that graph's own result.

\begin{table*}[t]
\centering
\small
\begin{tabular}{llccccc}
\toprule
Dataset & Method & Gemma2-2B & LLama3.2-3B & Falcon3-3B & Qwen3-4B & Average \\
\midrule

\multirow{8}{*}{GSM8k+}
& ICL with FS & 0.12\std{0.0} & 0.11\std{0.0} & 0.29\std{0.0} & 0.18\std{0.0} & 0.17\std{0.07} \\
& Tree-of-Thoughts & 0.00\std{0.00} & 0.01\std{0.00} & 0.02\std{0.00} & 0.12\std{0.00} & 0.04\std{0.05} \\
& Answer-only SFT & 0.00\std{0.00} & 0.00\std{0.00} & 0.00\std{0.00} & 0.00\std{0.00} & 0.00\std{0.00} \\
& CoT + Answer SFT & 0.08\std{0.00} & 0.24\std{0.00} & 0.26\std{0.00} & 0.37\std{0.01} & 0.24\std{0.10} \\
& Graph SFT & \textit{0.24}\std{0.01} & \textit{0.31}\std{0.00} & \textit{0.33}\std{0.00} & \textit{0.43}\std{0.00} & \textit{0.33}\std{0.07} \\
& Default GRPO & 0.00\std{0.00} & 0.03\std{0.00} & 0.04\std{0.00} & 0.24\std{0.01} & 0.08\std{0.09} \\
\rowcolor{lightgreen}
& Graph-Based GRPO & \textbf{0.27}\std{0.01} & \textbf{0.34}\std{0.01} & \textbf{0.36}\std{0.00} & \textbf{0.44}\std{0.01} & \textbf{0.35}\std{0.06} \\
\midrule

\multirow{8}{*}{iGSM}
& ICL with FS & 0.02\std{0.0} & 0.05\std{0.0} & 0.07\std{0.0} & 0.14\std{0.0} & 0.07\std{0.04} \\
& Tree-of-Thoughts & 0.29\std{0.00} & 0.22\std{0.00} & 0.29\std{0.00} & 0.49\std{0.00} & 0.32\std{0.10} \\
& Answer-only SFT & \textit{0.39}\std{0.01} & 0.45\std{0.01} & 0.44\std{0.01} & 0.49\std{0.01} & 0.44\std{0.04} \\
& CoT + Answer SFT & 0.21\std{0.01} & \textit{0.56}\std{0.01} & \textit{0.55}\std{0.01} & \textit{0.65}\std{0.0} & \textit{0.49}\std{0.17} \\
& Graph SFT & \textit{0.45}\std{0.00} & 0.45\std{0.00} & 0.43\std{0.00} & 0.55\std{0.00} & 0.47\std{0.05} \\
& Default GRPO & 0.36\std{0.01} & 0.17\std{0.03} & 0.28\std{0.01} & 0.45\std{0.01} & 0.32\std{0.10} \\
\rowcolor{lightgreen}
& Graph-Based GRPO & \textbf{0.57}\std{0.02} & \textbf{0.59}\std{0.00} & \textbf{0.56}\std{0.01} & \textbf{0.66}\std{0.01} & \textbf{0.59}\std{0.04} \\

\midrule
\multirow{8}{*}{MATH}
& ICL with FS & 0.03\std{0.0} & 0.08\std{0.0} & 0.12\std{0.0} & 0.20\std{0.0} & 0.11\std{0.06} \\
& Tree-of-Thoughts & \textit{0.25}\std{0.00} & \textit{0.51}\std{0.00} & \textit{0.52}\std{0.00} & \textit{0.76}\std{0.00} & \textit{0.51}\std{0.18} \\
& Answer-only SFT & 0.04\std{0.0} & 0.32\std{0.01} & 0.23\std{0.0} & \textit{0.43}\std{0.0} & 0.26\std{0.14} \\
& CoT + Answer SFT & \textit{0.10}\std{0.01} & \textit{0.51}\std{0.01} & \textit{0.40}\std{0.0} & \textit{0.59}\std{0.0} & \textit{0.40}\std{0.19} \\
& Graph SFT & 0.09\std{0.0} & 0.12\std{0.0} & 0.13\std{0.0} & 0.24\std{0.01} & 0.14\std{0.06} \\
\rowcolor{lightgreen}
& Default GRPO & \textbf{0.34}\std{0.01} & \textbf{0.63}\std{0.01} & \textbf{0.61}\std{0.01} & \textbf{0.85}\std{0.00} & \textbf{0.61}\std{0.18} \\
\rowcolor{lightred}
& Graph-Based GRPO & 0.12\std{0.00} & 0.17\std{0.01} & 0.17\std{0.01} & 0.30\std{0.00} & 0.19\std{0.07} \\

\midrule

\multirow{7}{*}{Entailment}
& ICL with FS & 0.00\std{0.00} & 0.00\std{0.00} & 0.00\std{0.00} & 0.11\std{0.00} & 0.03\std{0.05} \\
& Tree-of-Thoughts & 0.04\std{0.00} & 0.02\std{0.00} & 0.03\std{0.00} & 0.13\std{0.00} & 0.06\std{0.04} \\
& Answer-only SFT & 0.00\std{0.00} & 0.01\std{0.00} & 0.01\std{0.00} & 0.10\std{0.00} & 0.03\std{0.04} \\
& CoT + Answer SFT & 0.02\std{0.00} & 0.04\std{0.00} & 0.02\std{0.00} & 0.10\std{0.01} & 0.04\std{0.03} \\
& Graph SFT & \textit{0.05}\std{0.00} & \textit{0.07}\std{0.00} & \textit{0.07}\std{0.00} & \textit{0.17}\std{0.01} & \textit{0.09}\std{0.05} \\
& Default GRPO & 0.00\std{0.00} & 0.00\std{0.00} & 0.00\std{0.00} & 0.00\std{0.00} & 0.00\std{0.00} \\
\rowcolor{lightgreen}
& Graph-Based GRPO & \textbf{0.08}\std{0.03} & \textbf{0.10}\std{0.04} & \textbf{0.14}\std{0.02} & \textbf{0.20}\std{0.01} & \textbf{0.13}\std{0.05} \\

\midrule
\multirow{7}{*}{ProofWriter}
& ICL with FS & 0.05\std{0.00} & 0.07\std{0.00} & 0.25\std{0.00} & 0.40\std{0.00} & 0.19\std{0.16} \\
& Tree-of-Thoughts & 0.00\std{0.00} & 0.02\std{0.00} & 0.00\std{0.00} & 0.56\std{0.00} & 0.15\std{0.28} \\
& Answer-only SFT & 0.22\std{0.01} & 0.20\std{0.01} & 0.18\std{0.00} & 0.38\std{0.00} & 0.24\std{0.09} \\
& CoT + Answer SFT & \textit{0.44}\std{0.00} & 0.41\std{0.02} & \textit{0.36}\std{0.01} & 0.69\std{0.00} & 0.48\std{0.15} \\
& Graph SFT & 0.27\std{0.00} & \textit{0.54}\std{0.01} & 0.35\std{0.00} & \textit{0.77}\std{0.01} & \textit{0.48}\std{0.22} \\
& Default GRPO & 0.04\std{0.03} & 0.17\std{0.02} & 0.04\std{0.06} & 0.13\std{0.01} & 0.10\std{0.07} \\
\rowcolor{lightgreen}
& Graph-Based GRPO & \textbf{0.51}\std{0.00} & \textbf{0.58}\std{0.01} & \textbf{0.53}\std{0.02} & \textbf{0.81}\std{0.02} & \textbf{0.61}\std{0.14} \\

\bottomrule
\end{tabular}
\caption[Performance across datasets and methods]{%
Performance with standard deviations across datasets and training methods.
Best results are shown in bold and second-best in italic.}
\label{tab:main_results}
\end{table*}

\section{Experimental Setup}
\label{sec:setup}

\subsection{Models}
\label{subsec:models} We evaluate our approach using four instruction-tuned language models from four model families: Gemma 2 2B, Falcon 3 3B, Llama 3.2 3B, and Qwen3 4B.

\subsection{Datasets}
\label{subsec:data}

We evaluate on five benchmarks: three arithmetic reasoning tasks, where
solutions decompose into chains of operations, and two proof-construction
tasks, where the graph encodes an entailment structure rather than a
computation. Table~\ref{tab:datasets_stats} reports sizes and graph statistics.

\textbf{GSM8k+}~\citep{cobbe2021training} We build on GSM8k, $\sim$7K
grade-school word problems requiring multi-step arithmetic. Because the
benchmark is likely present in pre-training corpora
(Section~\ref{subsec:contamination}), test performance may partly reflect
memorization. We therefore keep each word problem and its computational graph
intact but redefine the four arithmetic operators: a script maps each of $+$,
$-$, $\times$, $/$ to a per-example linear combination of its two operands. We
re-execute the graph under the new semantics, recompute the gold answer, adapt
the natural-language solution, and state the operator definitions in the prompt
at training and inference time. The reasoning structure is unchanged, but the
memorized answer is now wrong (Appendix~\ref{app:gsm_rebuild}).

\textbf{iGSM}~\citep{Allen:23} $\sim$3.3K procedurally generated problems;
difficulty scales with the size of the underlying reasoning graph, and many
objects and relations are redundant.

\textbf{MATH}~\citep{hendrycks2021measuring} $\sim$10K high-school competition
problems spanning diverse topics and difficulty levels.

\textbf{Entailment}~\citep{dalvi2021entailmentbank} Expert-annotated
grade-school science QA requiring multi-step entailment proofs from facts to a
hypothesis. We use Task2, where relevant and distractor facts are both given
and most go unused; difficulty grows with proof length.

\textbf{ProofWriter}~\citep{tafjord2021proofwriter} True/False questions where
each label is justified by a proof over a theory of facts and rules, some of
which are distractors. The model must build a path from the relevant premises
to the queried statement.

\paragraph{Evaluation.}
For arithmetic datasets (GSM8K+, iGSM, and MATH), we execute
the generated graph $\hat{g}$ in topological order and take
the value of its final node as the prediction $\hat{y}$.
For free-form baselines, we extract the last number from
the generated response. We evaluate predictions using
exact match against the gold answer $y$.

For graph-building datasets (Entailment and ProofWriter),
we adopt the Overall AllCorrect metric \citep{dalvi2021entailmentbank}
from the official EntailmentBank evaluation framework.
This metric requires three components to be simultaneously
correct: the sets of leaf nodes and inference steps must
exactly match their gold counterparts, while intermediate
conclusions must achieve a BLEURT-F1 score of 1.
Overall AllCorrect is the product of these three binary
indicators, averaged across test examples.
We evaluate this metric on proof graphs from Graph SFT and Graph-Based GRPO, and on native proofs from the remaining methods.

For the internal/external faithfulness evaluation
(Appendix, Table~\ref{tab:faithfulness_full}),
we instead use answer-level exact match for ProofWriter,
independently of the full proof structure.
Internal accuracy compares the graph’s final-node result with the gold label; external accuracy compares the model’s explicit True/False answer with the gold label.

\paragraph{Splits.} Each dataset is split 80/20 into train and test (71/29 for
ProofWriter). For GRPO-Based training the training part is divided equally
between SFT and RL.

\subsection{Contamination Audit}
\label{subsec:contamination}
\begin{table*}[t]
\centering
\small
\setlength{\tabcolsep}{10pt}
\begin{tabular}{l c c c c c}
\toprule
Dataset & Qwen3-4B & Pythia-1.4B & Llama-3.2-3B-Instruct & Gemma-2-2B & Falcon3-3B-Instruct \\
\midrule
MATH         & 45.9 & 14.9 & 51.2 & 26.3 & 70.8 \\
GSM8K        & 28.2 & 1.3  & 34.9           & 21.2          & 78.1              \\
GSM8K+       & 0.0  & 0.0  & 0.0            & 0.0           & 0.0               \\
iGSM         & 0.0  & 0.0  & 0.0            & 0.0           & 1.3               \\
Entailment   & 9.6  & 0.0  & 0.0            & 0.0           & 0.3               \\
ProofWriter  & 0.3  & 0.0  & 0.0            & 0.0           & 31.1              \\
\bottomrule
\end{tabular}
\caption{CoDeC contamination scores (\%) computed for each benchmark dataset. The score is the
fraction of samples with a negative in-context delta $\Delta(x)<0$; lower values indicate less
contamination.}
\label{tab:codec_scores}

\end{table*}

Strong scores on a public benchmark may reflect pre-training exposure rather
than reasoning. We audit all five datasets with
CoDeC~\citep{zawalski2025detecting}, which flags a sample as memorized when
prepending other samples from the same dataset \emph{lowers} its likelihood;
the dataset score is the fraction of such samples. Against Pythia-1.4B, whose
corpus predates both benchmarks - MATH and GSM8k - Qwen3-4B shows scores of $45.9\%$ and $28.2\%$ --
three and twenty times the reference -- while iGSM, Entailment and ProofWriter
stay below $10\%$ (Table~\ref{tab:codec_scores}). This motivates GSM8k+,
and it bounds what MATH can show: much of that test set is solvable by recall,
which attenuates the measured effect of any reasoning method.

\subsection{Baselines}
\label{subsec:baselines}

We compare our method (Graph SFT + Graph GRPO, Section~\ref{sec:method})
against six baselines spanning three paradigms. All fine-tuned models use
LoRA~\citep{hu2022lora}; hyperparameters are in
Appendix~\ref{app:hyper}.

\textit{In-context learning.} \textbf{ICL with few-shot} prompts the base model
with two $(x, g, y)$ demonstrations in the graph format and executes the
generated $\hat{g}$. \textbf{Tree-of-Thoughts}~\citep{yao2023tree} is a
stronger zero-update baseline that explores a tree of intermediate steps with
branching, self-evaluation and backtracking at inference time.

\textit{Supervised fine-tuning.} \textbf{Answer-only SFT} trains on $(x, y)$
pairs and emits an answer with no intermediate reasoning. \textbf{CoT + Answer
SFT} trains on $(x, c, y)$ triples, where $c$ is the free-form reasoning trace
from the original dataset. \textbf{Graph SFT} trains on $(x, g, y)$ tuples in
our format; it shares our data and output format but omits the GRPO stage,
isolating the contribution of supervised graph training.

\textit{Reinforcement learning.} \textbf{Default GRPO} trains free-form
reasoning in \texttt{<think>} tags with rewards for answer correctness and
format compliance, representing the standard RL setup in recent reasoning
models.

\section{Results}
\label{sec:results}

\subsection{Main Results}
\label{subsec:results}

\begin{figure}[t]
\centering
\includegraphics[width=\columnwidth]{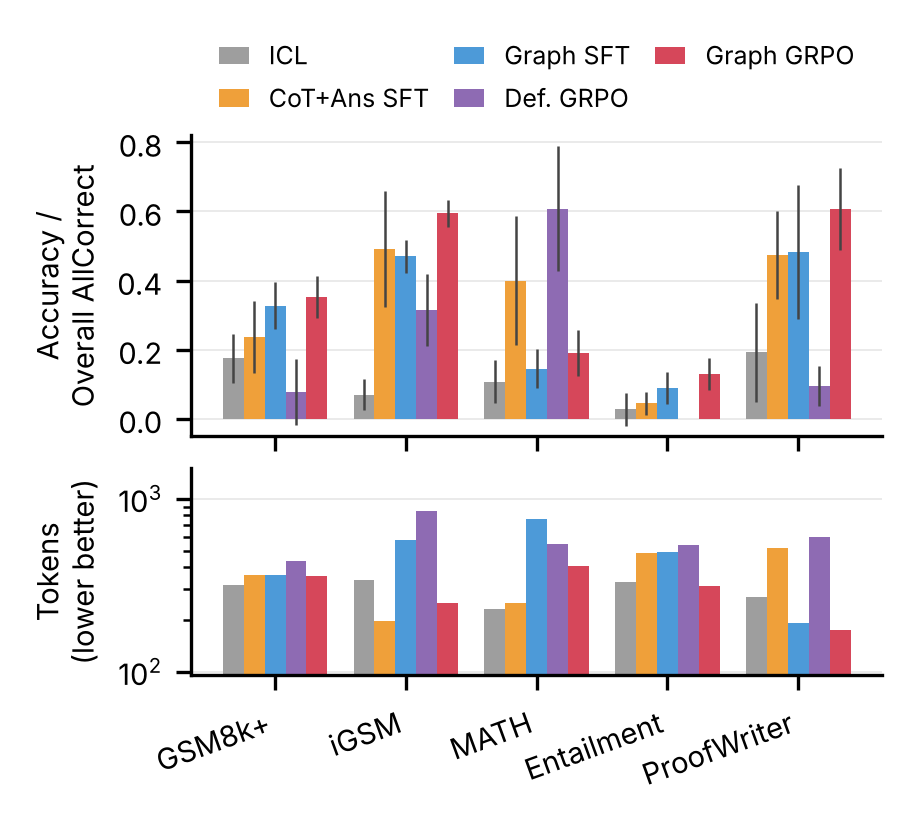}
\caption{Accuracy/Overall AllCorrect (top) and generation length (bottom, log scale), averaged
over the four models; error bars give the spread across models. Graph-Based
GRPO leads on four of five datasets at among the lowest token cost; MATH is the
exception. Tree-of-Thoughts and Answer-only SFT are omitted for scale and
because the latter emits no reasoning. Full results:
Tables~\ref{tab:main_results} and~\ref{tab:main_results_tokens}.}
\label{fig:performance_vs_tokens_barplot}
\end{figure}

Table~\ref{tab:main_results} and Figure~\ref{fig:performance_vs_tokens_barplot}
show that Graph-Based GRPO improves over Graph SFT on all five datasets,
with the strongest gains on compositional tasks. On GSM8k+ it reaches the best
average accuracy (0.35, against 0.33 for Graph SFT and 0.24 for CoT + Answer
SFT). The largest improvement is observed on ProofWriter, from 0.48 to 0.61, followed by iGSM, where accuracy increases from 0.47 to 0.59. Since iGSM is
procedurally generated to resist memorization, this suggests computational
graphs provide a useful inductive bias under distribution shift. On Entailment
and ProofWriter, Graph-Based GRPO is again best (0.13 and 0.61), consistent
with graph representations aligning naturally with multi-step proof
construction.

\begin{figure*}[t]
\centering
\includegraphics[width=\textwidth]{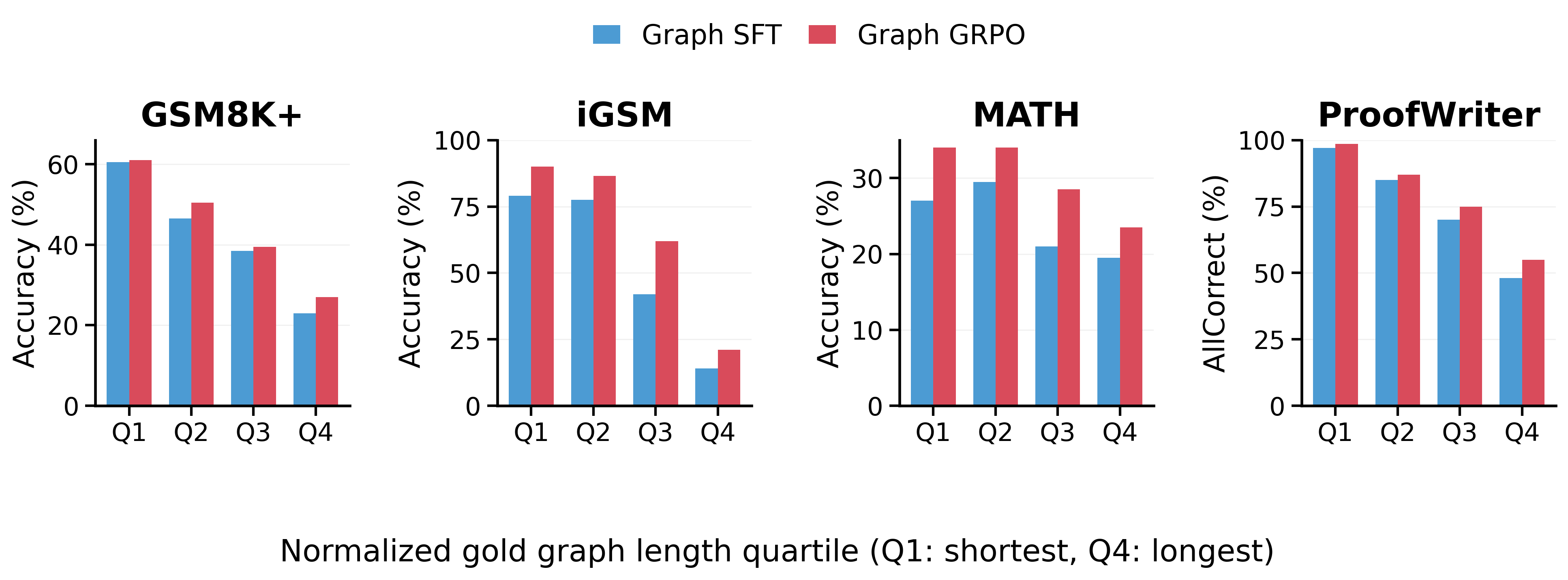}
\caption{Qwen3-4B accuracy (Overall AllCorrect on ProofWriter) stratified by normalized gold
graph length. Graph GRPO stays above Graph SFT in all bins and remains more
robust on longer graphs.}
\label{fig:acc_by_length}
\end{figure*}

MATH is the exception: Default GRPO (0.61), Tree-of-Thoughts (0.51), CoT + Answer SFT (0.40) and Answer-only SFT (0.26)
substantially outperform Graph-Based GRPO (0.19). Our contamination audit
(Section~\ref{subsec:contamination}) offers a partial explanation -- MATH is
the most contaminated dataset we test, so free-form methods can exploit
memorized natural-language solution patterns that structured reasoning does
not reproduce.

The lower panel of Figure~\ref{fig:performance_vs_tokens_barplot} shows
generation length. Graph-Based GRPO produces substantially shorter outputs than
Default GRPO while matching or exceeding its accuracy on four datasets; the
effect is largest on iGSM, where it is both the most accurate and the second
cheapest method. Compared with Graph SFT, Graph-Based GRPO generally produces shorter reasoning traces while achieving higher accuracy on all datasets (Appendix~\ref{app:token_consump}).

Per-model results (Table~\ref{tab:main_results}) show that
LLama3.2-3B, Falcon3-3B and Qwen3-4B benefit more consistently from graph-based
RL than Gemma2-2B, suggesting stable structured-reasoning policies require
sufficient model capacity.

Across models and datasets, most failures stem from incorrect operations or formulas, indicating that the main challenge is composing correct computations rather than generating graph structure. On GSM8k+, these errors account for 72–87\% of failures, while MATH and iGSM show greater variation and more structural errors in some settings. Smaller models, particularly Gemma2-2B, also produce more invalid graphs, highlighting difficulties with structural constraints. Overall, models generally generate valid graph structures but struggle with correct computational composition.More about graph errors at Appendix \ref{app:err_research}.

\subsection{Ablation}
\label{subsec:ablation}

\begin{table}[t]
\centering
\small
\setlength{\tabcolsep}{4pt}
\begin{tabular}{lcc}
\toprule
\textbf{Setting} & \textbf{Mean} & \textbf{Max} \\
\midrule
Full model & 43.88 $\pm$ 0.99 & 45.07 \\
\midrule
-- Style reward & 42.54 $\pm$ 0.89 & 43.79 \\
-- Correctness reward & 42.75 $\pm$ 0.65 & 43.44 \\
-- Overlap reward & 44.00 $\pm$ 1.12 & 45.49 \\
-- Default scheduler & 44.00 $\pm$ 0.86 & 45.21 \\
\midrule
1 rollout & 43.88 & 45.07 \\
8 rollouts + SC & 45.56 & 45.71 \\
16 rollouts + SC & \textbf{46.98} & \textbf{47.12} \\
\bottomrule
\end{tabular}
\caption{Ablation study of reward components, scheduling, and the number of rollouts for \textsc{Qwen3-4B} on GSM8k+. Mean accuracy is reported over seeds, with the best result in bold.}
\label{tab:ablation}
\end{table}

We ablate each component on Qwen3-4B using GSM8k+
(Table~\ref{tab:ablation}). Removing the style reward causes the largest drop
($-1.34$ points), and removing the correctness reward is nearly as damaging
($-1.13$). Removing the overlap reward slightly \emph{improves} the mean
($+0.12$), so on this benchmark structural agreement with the gold graph adds
little beyond the other two signals. 
Replacing the curriculum scheduler with the default scheduler yields
a small improvement ($+0.12$ percentage points). The curriculum
scheduler sorts training graphs by increasing number of nodes and
edges, training the model from simpler to more complex graphs,
whereas the default scheduler randomly shuffles training examples.
This suggests that curriculum ordering is not critical for final
accuracy in our setting.

Increasing the number of rollouts helps more than any reward change. Moving
from one rollout to eight with self-consistency~\citep{wang2023selfconsistencyimproveschainthought}
raises accuracy from $43.88\%$ to $45.56\%$, while sixteen rollouts further improve it to $46.98\%$, the best result in this ablation.

\subsection{Robustness to Graph Length and Sampling}
\label{subsec:robustness}

Figure~\ref{fig:acc_by_length} stratifies performance by gold graph length,
binned by percentile within each dataset. Across GSM8k+, iGSM, MATH and
ProofWriter both methods degrade as graphs grow, but Graph GRPO stays ahead in
every bin, with the clearest gaps at the upper percentiles on iGSM and
ProofWriter. RL fine-tuning therefore helps the model sustain multi-step graph
reasoning rather than only improving short instances.

On iGSM, Graph GRPO consistently outperforms Graph SFT in pass@$k$ and scales more reliably with additional samples: while SFT becomes increasingly variable across random seeds as $k$ grows, GRPO maintains stable performance (Appendix \ref{app:pass_eval}).

\section{Faithfulness}
\label{sec:faithfulness}

Accuracy alone cannot distinguish a model that derives its answer from the
generated graph from one that emits a plausible graph and answers by another
route. We assess this in two ways, summarized in
Table~\ref{tab:faithfulness}.

\paragraph{Internal--external agreement.}
We score each model twice: \emph{internally}, by executing the generated graph
and taking its final node, and \emph{externally}, by reading the answer the
model emits in its \texttt{ANSWER} field. A small gap $\Delta$ between the two
is necessary -- though not sufficient -- for the graph to be the source of
the answer.

On iGSM and MATH the gap nearly vanishes after GRPO ($1.0\!\rightarrow\!0.2$
and $1.7\!\rightarrow\!0.4$ points), and it falls by a factor of three on
ProofWriter. GSM8k+ is the exception: the gap is large for every method and
\emph{widens} under training, from $12.1$ for ICL to $31.9$ for Graph GRPO. The
cause is visible in the per-model numbers: for Qwen3-4B, internal accuracy is
$44.29\%$ while external accuracy against the redefined-operator gold is
$0.20\%$ -- yet external accuracy against the \emph{original} GSM8k answers is
$42.0\%$, almost matching the internal score. Despite the redefined operators
in the prompt, the model still emits the memorized answer, consistent with the
contamination audit in Section~\ref{subsec:contamination}.

\begin{table}[t]
\centering\footnotesize
\setlength{\tabcolsep}{2.5pt}
\begin{tabular}{@{}l ccccc@{}}
\toprule
Method & GSM8K+ & iGSM & MATH & PW \\
\midrule
\multicolumn{5}{@{}l}{\textit{$\Delta$ internal\,--\,external (pp, lower better)}} \\
ICL & \textbf{12.1}\sd{3} & 21.1\sd{18} & 25.2\sd{30} & 14.3\sd{5} \\
Graph SFT & 28.4\sd{9} & 1.0\sd{0} & 1.7\sd{1} & 11.8\sd{5} \\
\rowcolor{lightgreen}
Graph GRPO & 31.9\sd{9} & \textbf{0.2}\sd{0} & \textbf{0.4}\sd{0} & \textbf{3.5}\sd{1} \\
\midrule
\addlinespace[3pt]
\multicolumn{5}{@{}l}{\textit{$F_{\mathrm{strong}}$ (\%, higher better)}} \\
ICL & 0.0\sd{0} & 2.1\sd{1} & 1.7\sd{1} & 28.8\sd{25} \\
Graph SFT & 0.0\sd{0} & 44.0\sd{6} & 21.4\sd{6} & 49.4\sd{22} \\
\rowcolor{lightgreen}
Graph GRPO & 0.0\sd{0} & \textbf{48.2}\sd{3} & \textbf{40.2}\sd{2} & \textbf{52.4}\sd{22} \\
\bottomrule
\end{tabular}
\caption{Faithfulness, averaged over the four models; subscripts give the spread across models, rounded to the nearest point. $\Delta$ is the absolute gap between internal (graph execution) and external (emitted answer) accuracy; $F_{\mathrm{strong}}$ requires answer--graph consistency both before and after intervention. \textbf{Bold} = best per column. Entailment has no external answer; therefore,
neither internal--external accuracy nor
answer-based faithfulness metrics are applicable. Per-model values, and $F_{\mathrm{ID}}$/$F_{\mathrm{OOD}}$, are in Table~\ref{tab:faithfulness_full}.}
\label{tab:faithfulness}
\end{table}

\paragraph{Intervention-based faithfulness.}
Agreement is only correlational, so we additionally follow the intervention
protocol of \citet{somov2026breaking}, which treats the generated graph as a mediator $M$ with a deterministic executor $C(\cdot)$ mapping any graph to the answer it implies. $F_{\mathrm{ID}}$ asks whether the model's answer matches $C(\hat{M})$ for its own graph; $F_{\mathrm{OOD}}$ asks whether, after editing the graph
($\hat{M}\!\rightarrow\!M^{\star}$ by some kind of structural edit -- node or operation replacement) and
re-prompting, the new answer should also match $C(M^{\star})$; $F_{\mathrm{strong}}$
requires both $F_{\mathrm{ID}}$ and $F_{\mathrm{OOD}}$ to be true. We report $F_{\mathrm{strong}}$, the strictest of the three in Table \ref{tab:faithfulness}. 

Graph GRPO raises $F_{\mathrm{strong}}$ on every dataset where the metric is
meaningful: MATH nearly doubles ($21.4\!\rightarrow\!40.2$), and iGSM and ProofWriter improve from
already-high baselines ($44.0\!\rightarrow\!48.2$ and $49.4\!\rightarrow\!52.4$). Editing the graph therefore changes the answer more
reliably after RL than after supervised training alone.

GSM8k+ is again the outlier: $F_{\mathrm{strong}}$ is exactly $0$ for both
graph-based methods across all four models. Together with the widening internal–external gap, this suggests the model can construct a well-formed graph while relying on a separate memorized pathway for its answer.

\section{Conclusion}

We have explored structured reasoning as an alternative to free-form chain-of-thought by training language models to generate executable computational graphs. Our experiments show that reinforcement learning over graph-structured reasoning improves accuracy and efficiency, with Graph-Based GRPO consistently outperforming supervised graph training, particularly on compositional tasks, while producing shorter reasoning traces. Intervention experiments further reveal higher faithfulness, as modifying the reasoning graph changes the final answer, indicating genuine reliance on the generated structure. Overall, executable intermediate structures offer a promising path toward more reliable and verifiable reasoning in LLMs.

\section*{Limitations}

While our Graph-Based GRPO method demonstrates consistent improvements over Graph SFT, it still underperforms compared to CoT + Answer SFT and Default GRPO in some cases. This suggests that, despite the advantages of structured reasoning, there remains substantial room for improvement, particularly in the design of reward functions and output representations, as highlighted by our ablation study.

Our method is limited to tasks whose reasoning processes can be naturally represented as computational graphs. While this representation is suitable for mathematical reasoning and structured question answering, it may be less applicable to tasks that require open-ended, qualitative, or primarily linguistic reasoning that cannot be readily expressed as explicit computational dependencies.

\bibliography{custom}

\appendix
\section{Model's Hyperparameters and Training Methodology}
\label{app:hyper}
All models were trained with LoRA, 3 epochs for SFT pipelines, 600 iterations of GRPO for default GRPO pipeline and 1 epoch for pre-GRPO SFT checkpoint and 600 iterations for Graph GRPO pipeline. Models trained for 3 seed runs - 0, 42, 2026.

On average, SFT stages (answer-only, CoT+answer, and Graph SFT) take about 20 minutes per seed, while GRPO stages (default GRPO and Graph GRPO) take about 2.5 hours per seed. All runs used 4×A100 GPUs with 80 GB memory.

\paragraph{LoRA} $\alpha = 32, r = 32$, attention (q\_proj, k\_proj, v\_proj, o\_proj) and mlp (gate\_proj, up\_proj, down\_proj) modules are tuned.

\paragraph{GRPO} num\_generations=$16$ for Default GRPO and for Graph GRPO pipelines.

\paragraph{Other} lr=$5 \times 10^{-6}$ for SFT and GRPO pipelines, lr=$3 \times 10^{-4}$ for pre-GRPO checkpoint. paged\_adamw\_8bit optimizer \citep{kingma2014adam}, batch\_size=$2$, gradient\_accumulation=$2$.
\section{Computational graph generation algorithm}
\label{app:comp_gen}
To generate computational graphs for GSM8K+, iGSM and MATH, we use a few-shot prompting strategy. Specifically, two example problems from the same domain are included in the system prompt to demonstrate the desired graph format and guide the model toward producing structured outputs.

We use a low sampling temperature ($T = 0.2$) to ensure stable graph generation and reduce variability in the produced structures. All graphs are generated using GPT-4o.

For arithmetical datasets, to ensure the validity of the generated graphs, we execute a forward pass through each graph and compare the computed answer with the ground-truth solution. We additionally enforce that the final node of the graph must be an operation node rather than a constant node. This constraint prevents degenerate cases in which the model outputs the correct answer directly while the preceding nodes and edges do not contribute to the final computation.
The graph generation algorithm is identical for all three mathematical datasets.

For the graph-building datasets, Entailment and ProofWriter, we reconstruct computational graphs from the original solutions.

The prompt is in Appendix \ref{app:promt_for_graph}.

\section{GSM8k+ building} \label{app:gsm_rebuild}
\subsection{Motivation}
GSM8k is a standard benchmark, but strong scores on the public test set may reflect pre-training exposure rather than genuine reasoning. We therefore reconstruct the benchmark to reduce the risk that a model simply recalls familiar question--answer pairs from the original release.

\subsection{What we kept and what we changed}
We keep the original word problem and the same graph structure: the same nodes and the same assignment lines (e.g., \texttt{ryan=karl+20}, \texttt{ben=ryan-10}). Only the semantics of the four operators, \texttt{+}, \texttt{-}, \texttt{*}, and \texttt{/}, are changed. Using a script, each example receives its own mapping from these symbols to linear combinations of the two operands (e.g., \texttt{+} may be defined as $1\cdot a + 1\cdot b - 2$). The graph is then evaluated with these redefined operations, and the resulting value is used as the graph-level gold answer. We also modify our NLU solutions to account for the redefined operations. In addition, we add the new operator definitions to each prompt during both training and inference.

\subsection{Why this helps against contamination}
Because the underlying data structure is changed, a model cannot solve the task by memorizing the numbers from the original benchmark.

%

\section{Contamination Analysis of the Benchmark Datasets}
\label{app:contamination}

A reliable comparison of different reasoning strategies in LLMs requires verifying that the
benchmark datasets used for evaluation have not leaked into the models' pretraining corpora.
Data contamination inflates the reported metrics and confounds the measured effect of the
reasoning procedure itself, because a model that has memorized the evaluation data solves the
tasks by retrieving memorized patterns rather than by generalizing through the reasoning
process. We therefore perform a systematic contamination audit of all datasets considered in
this work.

\subsection{Detection Method}

We detect contamination using CoDeC (Contamination Detection via Context)~\cite{zawalski2025detecting},
a model- and dataset-agnostic method that requires only gray-box access to token probabilities.
For each sample $x$ in a candidate dataset $\mathcal{D}$, CoDeC compares two quantities:
(i) the \emph{baseline confidence}, defined as the average log-likelihood of the tokens of $x$
scored in isolation, and (ii) the \emph{in-context confidence}, obtained by prepending $n$
other samples drawn from $\mathcal{D}\setminus\{x\}$ to $x$ and measuring the average
log-likelihood of the tokens of $x$ in this augmented context. The contamination signal of a
sample is the difference
\begin{equation}
  \Delta(x) \;=\; \log p_{\mathrm{ic}}(x) \;-\; \log p_{\mathrm{base}}(x),
  \label{eq:delta}
\end{equation}
averaged over several context draws. For unseen data, additional in-context examples typically
increase the model's confidence ($\Delta(x)>0$), whereas for memorized data the context
disrupts memorization and reduces confidence ($\Delta(x)<0$). The dataset-level contamination
score is defined as the fraction of samples with $\Delta(x)<0$.

Following the recommendations of~\cite{zawalski2025detecting}, we use a single in-context example
($n=1$), average the per-sample delta over $5$ random context draws, and exclude the first
$10$ predicted tokens of the target when computing the average log-likelihood; input sequences
are truncated to at most $4096$ tokens.

\subsection{Contamination Scores}

Table~\ref{tab:codec_scores} reports the CoDeC scores computed for four models--Qwen3-4B,
Llama-3.2-3B-Instruct, Gemma-2-2B, and Falcon3-3B-Instruct--together with a reference model,
Pythia-1.4B (Section~\ref{app:stat_tests}). For MATH and GSM8K the fraction of samples with a
negative delta is elevated across all four models, ranging from $26.3\%$ (Gemma-2-2B) to
$70.8\%$ (Falcon3-3B-Instruct) on MATH and from $21.2\%$ (Gemma-2-2B) to $78.1\%$
(Falcon3-3B-Instruct) on GSM8K. By contrast, the scores for iGSM and Entailment remain below
$10\%$ for all models. For ProofWriter, the scores are near zero for Qwen3-4B,
Llama-3.2-3B-Instruct, and Gemma-2-2B, whereas Falcon3-3B-Instruct reaches $31.1\%$.

We also audit the redefined-operator variant GSM8K+. This dataset is derived from the exact
same underlying problems as the original benchmark: the wording, numbers, format, and difficulty
are unchanged, and only the operator definitions (and the resulting solutions) are replaced, so
that the resulting texts could not have appeared in any pretraining corpus. The contamination
score of GSM8K+ is $0.0\%$ for every model (Table~\ref{tab:codec_scores}), with
strictly positive per-sample deltas (mean $\Delta\approx+1.1$ to $+1.4$). Because the redefined
version holds every property of the original benchmark constant except the specific, memorizable
text, its zero score rules out any explanation of the original benchmark's elevated scores based
on format, wording, or task difficulty. The only remaining explanation for the discrepancy is that
the specific text of the original GSM8K was memorized, which provides an additional, within-model
confirmation that this benchmark partially leaked into the models' pretraining data.

\subsection{Statistical Comparison against a Reference Model}
\label{app:stat_tests}

To establish that the elevated scores of MATH and GSM8K reflect genuine memorization rather
than dataset-specific properties (e.g., an intrinsically low baseline confidence), we follow
the reference-based comparison recommended by~\cite{zawalski2025detecting} and contrast the scores
of each target model with those of a reference model known to be uncontaminated. We use
Pythia-1.4B, which was pretrained on The Pile, a corpus whose composition was frozen in 2021,
before the release of MATH and GSM8K; neither benchmark could therefore appear in its training
data.

For each pair of per-sample $\Delta$ distributions we conduct two one-sided tests. The
\emph{Mann--Whitney $U$ test} (Wilcoxon rank-sum test) is applied to the per-sample $\Delta$
values, with the alternative hypothesis that the deltas of the target model are stochastically
smaller (more negative) than those of the reference; the magnitude of the effect is quantified
by Cliff's $\delta$. The \emph{Fisher's exact test} is applied to the $2\times2$ contingency
table of negative ($\Delta<0$) versus non-negative ($\Delta\ge0$) samples, with the alternative
hypothesis that the odds of a negative delta are higher in the target model than in the
reference. We additionally report 95\% Wilson score intervals for both contamination
proportions.

Table~\ref{tab:stat_tests} summarizes the results for MATH and GSM8K.

\begin{table*}[t]
\centering
\footnotesize
\setlength{\tabcolsep}{4pt}
\begin{tabular}{@{}l r@{\,/\,}l r@{\,/\,}l ccc cc@{}}
\toprule
& \multicolumn{2}{c}{Score (\%)} & \multicolumn{2}{c}{Wilson 95\% CI (\%)}
& \multicolumn{3}{c}{Mann--Whitney $U$} & \multicolumn{2}{c}{Fisher's exact} \\
\cmidrule(lr){2-3} \cmidrule(lr){4-5} \cmidrule(lr){6-8} \cmidrule(l){9-10}
Model & \multicolumn{2}{c}{Model\,/\,Pythia} & \multicolumn{2}{c}{Model\,/\,Pythia}
& $U$ & $p$ & Cliff's $\delta$ & OR & $p$ \\
\midrule
\multicolumn{10}{@{}l}{\textit{MATH}} \\
Qwen3-4B              & 45.9 & 14.9 & 43.7--48.1 & 13.4--16.6 & 1\,403\,460    & $<10^{-6}$ & 0.276 & 4.831   & $<10^{-6}$ \\
Llama-3.2-3B-Instruct & 51.2 & 14.8 & 50.1--52.3 & 14.1--15.6 & 15\,448\,768.5 & $<10^{-6}$ & 0.498 & 6.016   & $<10^{-6}$ \\
Gemma-2-2B            & 26.3 & 14.8 & 25.3--27.3 & 14.1--15.6 & 17\,161\,310   & $<10^{-6}$ & 0.443 & 2.045   & $<10^{-6}$ \\
Falcon3-3B-Instruct   & 70.8 & 14.8 & 69.8--71.8 & 14.1--15.6 & 10\,488\,688.5 & $<10^{-6}$ & 0.660 & 13.901  & $<10^{-6}$ \\
\addlinespace[3pt]
\multicolumn{10}{@{}l}{\textit{GSM8K}} \\
Qwen3-4B              & 28.2 & 1.3  & 25.9--30.6 & 0.8--2.0   & 940\,049       & $2.5\times10^{-3}$ & 0.061 & 30.479  & $<10^{-6}$ \\
Llama-3.2-3B-Instruct & 34.9 & 1.3  & 32.5--37.4 & 0.8--2.0   & 168\,582.5     & $<10^{-6}$ & 0.832 & 41.629  & $<10^{-6}$ \\
Gemma-2-2B            & 21.2 & 1.3  & 19.2--23.4 & 0.8--2.0   & 191\,762.5     & $<10^{-6}$ & 0.809 & 20.882  & $<10^{-6}$ \\
Falcon3-3B-Instruct   & 78.1 & 1.3  & 75.9--80.2 & 0.8--2.0   & 53\,643        & $<10^{-6}$ & 0.946 & 276.646 & $<10^{-6}$ \\
\bottomrule
\end{tabular}
\caption{Two-sample comparison of the per-sample CoDeC deltas between each target model and
Pythia-1.4B (reference). Values are reported as target\,/\,reference. The Mann--Whitney $U$ test
is one-sided ($H_1$: target deltas stochastically smaller) and Fisher's exact test is one-sided
($H_1$: higher odds of a negative delta in the target); Cliff's $\delta$ is the effect size for
the rank test.}
\label{tab:stat_tests}
\end{table*}

\subsection{Comparison of Original and Redefined-Operator Datasets}
\label{app:redefined_vs_orig}

To quantify this additional evidence of memorization, we contrast the original GSM8K with its
redefined-operator counterpart GSM8K+ evaluated on the same model. Table~\ref{tab:redefined_vs_orig}
reports the two-sample comparison between the original dataset (target) and the redefined
dataset (reference). For every model the original dataset has a strictly lower per-sample delta
distribution than the redefined dataset: the Mann--Whitney $U$ test is significant at
$p<10^{-6}$ in all cases, Cliff's $\delta$ ranges from $0.97$ to $1.00$, and Fisher's exact test
yields an infinite odds ratio because the redefined dataset contains no negative deltas at all.
The difference between the original and the redefined dataset is therefore not only large but
also statistically unambiguous.

\begin{table*}[t]
\centering
\caption{Two-sample comparison of the original GSM8K versus its redefined-operator version
(GSM8K+) on the same model. Values are reported as original~/~redefined. The Mann--Whitney
$U$ test is one-sided ($H_1$: original deltas stochastically
smaller) and Fisher's exact test is one-sided ($H_1$: higher odds of a negative delta in the
original); Cliff's $\delta$ is the effect size for the rank test.}
\label{tab:redefined_vs_orig}
\small
\setlength{\tabcolsep}{5pt}
\begin{tabular}{ll r@{\,/}l r@{\,/}l c c c c c}
\toprule
& & \multicolumn{2}{c}{Score (\%)} &
\multicolumn{2}{c}{Wilson 95\% CI (\%)} & \multicolumn{3}{c}{Mann--Whitney $U$} &
\multicolumn{2}{c}{Fisher's exact} \\
\cmidrule(lr){3-4} \cmidrule(lr){5-6}
\cmidrule(lr){7-9} \cmidrule(lr){10-11}
Model & Dataset & \multicolumn{2}{c}{Orig.~/~Redef.} &
\multicolumn{2}{c}{Orig.~/~Redef.} & $U$ & $p$ & Cliff's $\delta$ & OR & $p$ \\
\midrule
Qwen3-4B              & GSM8K & 28.2 & 0.0 & 25.9--30.6 & 0.0--0.3 & 25\,530.0 & $<10^{-6}$ & 0.974 & $\infty$ & $<10^{-6}$ \\
Llama-3.2-3B-Instruct & GSM8K & 34.9 & 0.0 & 32.5--37.4 & 0.0--0.3 & 0.0 & $<10^{-6}$ & 1.000 & $\infty$ & $<10^{-6}$ \\
Gemma-2-2B            & GSM8K & 21.2 & 0.0 & 19.2--23.4 & 0.0--0.3 & 0.0 & $<10^{-6}$ & 1.000 & $\infty$ & $<10^{-6}$ \\
Falcon3-3B-Instruct   & GSM8K & 78.1 & 0.0 & 75.9--80.2 & 0.0--0.3 & 0.0 & $<10^{-6}$ & 1.000 & $\infty$ & $<10^{-6}$ \\
\bottomrule
\end{tabular}
\end{table*}

\subsection{Discussion}

For MATH, the contamination scores of all four target models exceed that of Pythia-1.4B
($\approx14.8\%$), and the differences are statistically significant (Mann--Whitney $p<10^{-6}$;
Fisher's exact $p<10^{-6}$): Qwen3-4B reaches $45.9\%$, Llama-3.2-3B-Instruct $51.2\%$,
Gemma-2-2B $26.3\%$, and Falcon3-3B-Instruct $70.8\%$. For GSM8K the discrepancy is even more
pronounced: all target models score far above Pythia-1.4B ($1.3\%$), with Falcon3-3B-Instruct
reaching $78.1\%$ and Fisher's odds ratios of up to $\approx 277$. These results provide strong
evidence that MATH and GSM8K were partially present in the pretraining data of all four models,
most severely in Falcon3-3B-Instruct. In addition, Falcon3-3B-Instruct shows a notable
ProofWriter score ($31.1\%$), suggesting partial contamination of that dataset as well.

This contamination may explain why the reasoning approaches introduced in this paper yield
comparatively modest improvements on MATH and GSM8K: a substantial fraction of the evaluation
examples are solved through memorization rather than through the reasoning process, which
attenuates the measured effect of the proposed methods. Finally, the clean $0.0\%$ score of
GSM8K+ across all models reinforces this conclusion: this dataset
retains the exact wording, format, and difficulty of the original benchmark and differs only in
the redefined operators, yet it produces no contamination signal at all. The fact that a
minimal, non-memorizable textual change is sufficient to erase the contamination signal shows
that the elevated scores of the original GSM8K are due to memorization of its
specific text, providing further evidence that it partially leaked into the models'
pretraining data.

\begin{table*}[t]
\centering
\small
\begin{tabular}{llcccc}
\toprule
Dataset & Method & Gemma2-2B & LLama3.2-3B & Falcon3-3B & Qwen3-4B \\
\midrule
\multirow{7}{*}{GSM8k+}
& ICL with FS
& 105\std{46} & 188\std{173} & 86\std{40} & 880\std{312} \\
& Tree-of-Thoughts
& 2476\std{396} & 4442\std{594} & 4868\std{775} & 6907\std{1031} \\
& Answer-only SFT
& 8\std{2} & 4\std{10} & 12\std{86} & 3\std{1} \\
& CoT + Answer SFT
& 869\std{213} & 112\std{92} & 251\std{149} & 216\std{130} \\
& Graph SFT
& 348\std{4} & 267\std{2} & 401\std{7} & 428\std{27} \\
& Default GRPO
& 216\std{4} & 364\std{28} & 504\std{35} & 669\std{89} \\
& Graph-Based GRPO
& 324\std{3} & 264\std{3} & 359\std{2} & 478\std{259} \\
\midrule
\multirow{7}{*}{iGSM}
& ICL with FS
& 326\std{118} & 361\std{161} & 218\std{136} & 445\std{137} \\
& Tree-of-Thoughts
& 2476\std{414} & 5072\std{740} & 4632\std{1220} & 5955\std{2109} \\
& Answer-only SFT
& 5\std{1} & 4\std{1} & 5\std{0} & 5\std{1} \\
& CoT + Answer SFT
& 410\std{167} & 127\std{145} & 131\std{150} & 112\std{122} \\
& Graph SFT
& 829\std{30} & 448\std{20} & 788\std{17} & 248\std{25} \\
& Default GRPO
& 317\std{47} & 832\std{23} & 840\std{18} & 1423\std{15} \\
& Graph-Based GRPO
& 303\std{23} & 216\std{15} & 251\std{9} & 233\std{16} \\
\midrule
\multirow{7}{*}{MATH}
& ICL with FS
& 279\std{124} & 165\std{177} & 92\std{90} & 383\std{168} \\
& Tree-of-Thoughts
& 2476\std{458} & 3817\std{1068} & 3878\std{1436} & 5213\std{2034} \\
& Answer-only SFT
& 5\std{6} & 4\std{0} & 6\std{6} & 3\std{1} \\
& CoT + Answer SFT
& 507\std{63} & 150\std{130} & 189\std{157} & 154\std{137} \\
& Graph SFT
& 823\std{68} & 580\std{7} & 1000\std{13} & 634\std{19} \\
& Default GRPO
& 349\std{13} & 404\std{5} & 633\std{24} & 795\std{4} \\
& Graph-Based GRPO
& 341\std{13} & 422\std{26} & 498\std{53} & 365\std{7} \\
\midrule
\multirow{7}{*}{Entailment}
& ICL with FS
& 372\std{88} & 325\std{72} & 156\std{50} & 458\std{51} \\
& Tree-of-Thoughts
& 2754\std{597} & 4588\std{587} & 4277\std{822} & 4134\std{606} \\
& Answer-only SFT
& 132\std{219} & 105\std{223} & 113\std{159} & 118\std{231} \\
& CoT + Answer SFT
& 712\std{395} & 358\std{377} & 572\std{419} & 306\std{319} \\
& Graph SFT
& 479\std{13} & 502\std{30} & 499\std{25} & 482\std{9} \\
& Default GRPO
& 194\std{34} & 786\std{0} & 660\std{5} & 517\std{169} \\
& Graph-Based GRPO
& 386\std{127} & 305\std{66} & 297\std{19} & 259\std{18} \\
\midrule
\multirow{7}{*}{ProofWriter}
& ICL with FS
& 163\std{46} & 237\std{61} & 551\std{179} & 128\std{39} \\
& Tree-of-Thoughts
& 2073\std{394} & 4132\std{520} & 2717\std{447} & 5242\std{677} \\
& Answer-only SFT
& 29\std{1} & 27\std{1} & 31\std{2} & 34\std{2} \\
& CoT + Answer SFT
& 459\std{11} & 846\std{8} & 425\std{19} & 354\std{5} \\
& Graph SFT
& 225\std{24} & 233\std{4} & 158\std{4} & 146\std{1} \\
& Default GRPO
& 446\std{90} & 727\std{123} & 694\std{235} & 533\std{96} \\
& Graph-Based GRPO
& 205\std{37} & 144\std{19} & 221\std{26} & 124\std{5} \\
\bottomrule
\end{tabular}
\caption[Token consumption across datasets and methods]{%
Token consumption comparison per generation across datasets and training methods.
Standard deviations are shown in smaller font for readability.}
\label{tab:main_results_tokens}
\end{table*}

\begin{table*}[t]
\centering
\small
\begin{tabular}{llcccc}
\toprule
Dataset & Model & Not valid graphs & Not initialized nodes & Unused nodes & Incorrect ops/formulas \\
\midrule

\multirow{4}{*}{GSM8k+}
& Falcon3-3B  & 2\std{0} & 2\std{0} & 18\std{2} & 78\std{2} \\
& Gemma2-2B   & 7\std{1} & 2\std{0} & 20\std{2} & 72\std{1} \\
& LLama3.2-3B & 4\std{1} & 1\std{0} & 19\std{3} & 75\std{2} \\
& Qwen3-4B    & 2\std{0} & 1\std{0} & 9\std{1} & 87\std{1} \\

\midrule

\multirow{4}{*}{iGSM}
& Falcon3-3B  & 1\std{0} & 2\std{1} & 13\std{4} & 85\std{3} \\
& Gemma2-2B   & 94\std{3} & 2\std{2} & 2\std{1} & 2\std{2} \\
& LLama3.2-3B & 1\std{0} & 2\std{1} & 26\std{9} & 71\std{10} \\
& Qwen3-4B    & 2\std{1} & 2\std{1} & 36\std{3} & 60\std{3} \\

\midrule

\multirow{4}{*}{MATH}
& Falcon3-3B  & 1\std{0} & 5\std{0} & 18\std{2} & 76\std{2} \\
& Gemma2-2B   & 52\std{32} & 13\std{7} & 10\std{5} & 26\std{33} \\
& LLama3.2-3B & 2\std{1} & 3\std{0} & 23\std{1} & 71\std{1} \\
& Qwen3-4B    & 3\std{1} & 5\std{0} & 19\std{2} & 73\std{2} \\

\bottomrule
\end{tabular}
\caption{Distribution of error categories (\%) for Graph-Based GRPO across datasets and models (mean and standard deviation over 3 seeds). Incorrectly initialized constant nodes are counted under \textit{Incorrect ops/formulas}.}
\label{tab:error-categories-percent-only-no-const}
\end{table*}

\section{Token consumption}
\label{app:token_consump}

Table~\ref{tab:main_results_tokens} reports the average number of tokens generated per solution across datasets, models, and training strategies. This analysis provides additional insight into the efficiency of different reasoning approaches.

\paragraph{Natural language reasoning vs structured reasoning.}
Graph-Based GRPO produces shorter outputs than Default GRPO in almost every setting. The gap is modest on GSM8k+ and MATH but grows to more than threefold on iGSM and ProofWriter, the tasks with the longest reasoning chains, and reaches about $6\times$ for Qwen3-4B on iGSM. Default GRPO rewards only answer correctness and format, so its free-form traces tend to grow during training. Graph-Based GRPO writes each step as a compact node with explicit inputs, which leaves little room for verbosity. These savings come with higher accuracy on four of the five datasets. Two out of three cases where Default GRPO is shorter coincide with near-zero accuracy. The advantage also extends beyond arithmetic. On the proof-construction tasks, Entailment and ProofWriter, Graph-Based GRPO is both shorter and more accurate than CoT + Answer SFT for every model.

\paragraph{Effect of reinforcement learning on generation length.}
One of the most notable trends in Table~\ref{tab:main_results_tokens} is that Graph-Based GRPO often reduces generation length relative to Graph SFT, although the effect depends on the model and dataset. Substantial reductions occur on iGSM, particularly for Falcon3-3B (788 to 251 tokens) and Gemma2-2B (829 to 303 tokens). Similar reductions can be observed for Qwen3-4B on iGSM (248 → 233 tokens) and MATH (634 → 365 tokens). However, generation length increases in some settings, including Qwen3-4B on GSM8k+ (428 to 478 tokens) and Falcon3-3B on ProofWriter (158 to 221 tokens).

This suggests that reinforcement learning not only improves reasoning accuracy but also encourages the model to generate more concise computational graphs. Intuitively, GRPO rewards solutions that lead to correct answers, which implicitly discourages unnecessary intermediate nodes and redundant operations.

\paragraph{Comparison with Default GRPO.}
The difference between Default GRPO and Graph-Based GRPO is particularly striking in terms of token efficiency. Default GRPO often produces very long outputs, especially on more difficult datasets such as iGSM. For example, Default GRPO with Falcon3-3B generates on average 840 tokens on iGSM, compared to only 251 tokens for Graph-Based GRPO. A similar pattern can be observed for Qwen3-4B (1423 vs 233 tokens).

These results indicate that graph-based reasoning provides a much more compact representation of the reasoning process than natural language chain-of-thought. Instead of describing intermediate steps in natural language, the model directly produces executable reasoning graphs.

\paragraph{Gemma token anomaly.}
An interesting anomaly can be observed for the Gemma2-2B model. Despite being the smallest model in our study (2B parameters), it often produces unusually long outputs, particularly when trained with CoT-based methods. For example, on GSM8k+ the CoT + Answer SFT approach generates more than 860 tokens on average, significantly exceeding the generation length of larger models such as LLama3.2-3B or Falcon3-3B.

Manual inspection suggests that this behavior is frequently associated with hallucinated reasoning traces, where the model generates excessively verbose or redundant intermediate steps that do not contribute to the final computation. This phenomenon is consistent with previous observations that smaller models tend to rely on longer and less structured reasoning traces when attempting to solve complex tasks.

\paragraph{Summary.}
Overall, the results show that graph-based reasoning significantly reduces token consumption while maintaining or improving performance. Reinforcement learning further enhances this effect by encouraging more compact reasoning graphs. These findings suggest that computational graphs provide a more efficient intermediate representation for reasoning compared to natural language explanations.

\section{Multi-seed pass@$k$ evaluation.}
\label{app:pass_eval}
\begin{figure}[H]
    \centering
    \includegraphics[width=\linewidth]{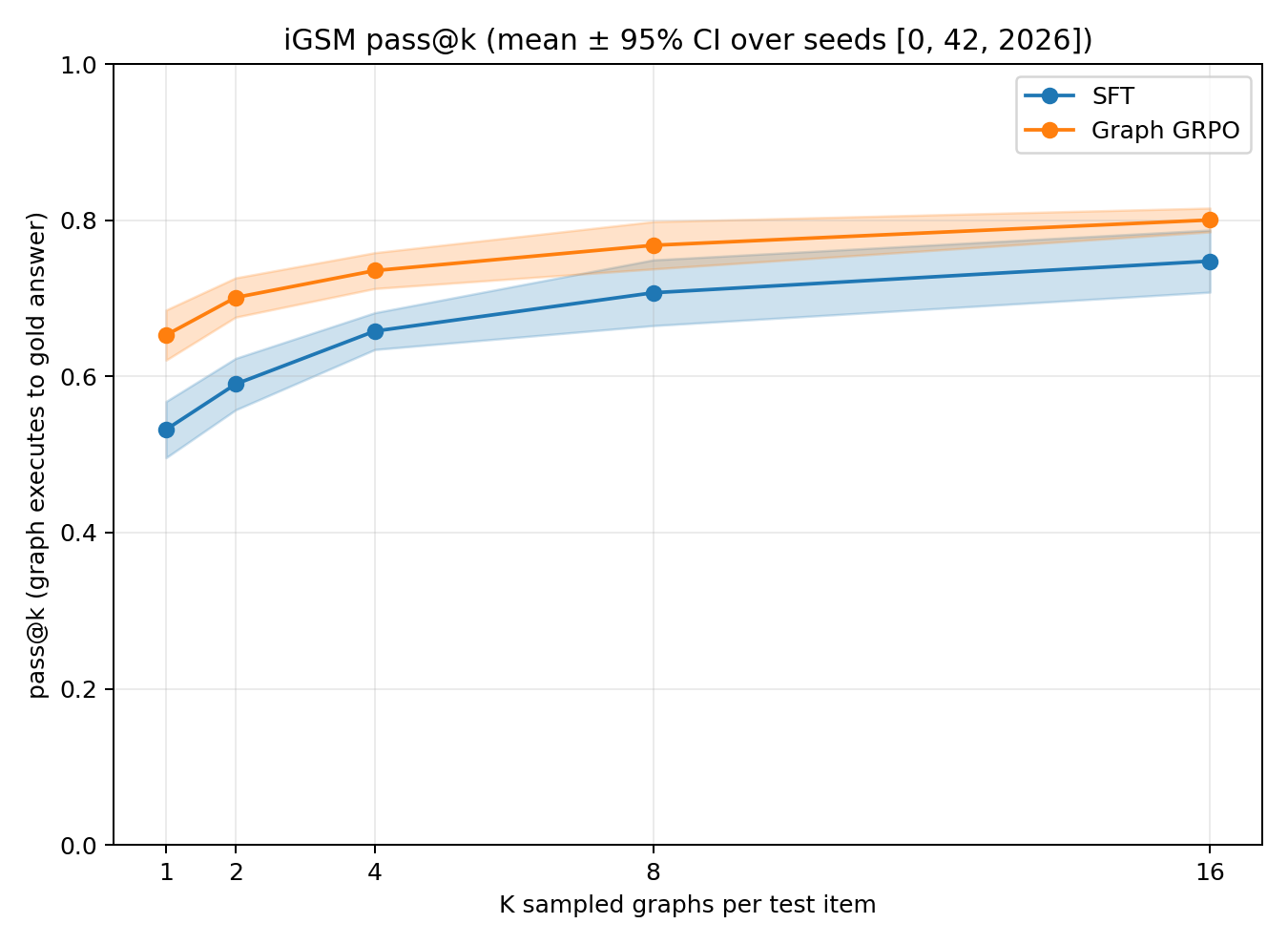}
    \caption{Pass@$k$ on iGSM across three random seeds. Shaded regions indicate 95\% confidence intervals over seeds. Graph GRPO consistently outperforms SFT and exhibits more stable performance as the number of sampled graphs increases.}
    \label{fig:pass_at_k_multiseed}
\end{figure}

Figure~\ref{fig:pass_at_k_multiseed} reports pass@$k$ on iGSM across three random seeds, with shaded regions denoting 95\% confidence intervals. Graph GRPO consistently outperforms Graph SFT across all $k$ values. Moreover, its confidence interval remains relatively stable as $k$ increases, whereas SFT exhibits greater seed-level variability. This suggests that Graph GRPO benefits from additional samples without substantially increasing sensitivity to the training seed.

\section{Error research}
\label{app:err_research}

Table~\ref{tab:error-categories-percent-only-no-const} presents a detailed breakdown of error categories across datasets and models.
Several trends can be observed from the table.

\paragraph{Dominance of reasoning errors.}
Across most models and datasets, the largest proportion of failures is caused by incorrect operations or formulas. This indicates that the primary difficulty lies in selecting and composing the correct arithmetic operations rather than generating the graph structure itself.

\paragraph{Dataset difficulty effects.}
The distribution of error types varies across datasets.
On GSM8k+, incorrect operations and formulas dominate,
accounting for 72--87\% of errors. A similar pattern is
observed on MATH and iGSM, although these datasets exhibit
greater variation in error distributions, with a higher
proportion of structural errors in some settings.

\paragraph{Model size effects.}
Smaller models struggle more with producing valid structured outputs. In particular, Gemma2-2B exhibits a very high rate of invalid graphs on iGSM and MATH, suggesting that maintaining strict structural constraints is difficult for smaller models.

\paragraph{Unused and missing nodes.}
Errors related to unused nodes and missing node initialization occur when the model partially decomposes the problem but fails to correctly connect intermediate steps. These errors highlight the difficulty of maintaining consistent dependency structures within the generated graphs.

Overall, these results suggest that while models are generally capable of generating the correct graph format, the main challenge remains learning correct computational compositions within the graph structure.

\clearpage
\onecolumn
\section{Faithfulness: Full Results}
\label{app:faithfulness_full}
\begin{center}
\small
\setlength{\tabcolsep}{2pt}
\begin{tabular}{llcccc}
\toprule
Dataset & Method & Gemma2-2B & LLama3.2-3B & Falcon3-3B & Qwen3-4B \\
\midrule
\multicolumn{6}{l}{\textit{(a) Internal (graph execution) / External (emitted answer) accuracy (\%).}} \\
\midrule
\multirow{3}{*}{GSM8K+} & ICL
& 9.23/\textbf{0.14}
& 10.93/0.07
& 19.02/\textbf{1.06}
& 10.79/\textbf{21.22} \\
 & Graph SFT & 19.33/0.09 & 25.34/\textbf{0.38} & 27.06/0.26 & 43.20/0.51 \\
 & Graph-Based GRPO & \textbf{19.92}/\textbf{0.14} & \textbf{29.88}/0.24 & \textbf{34.47}/0.31 & \textbf{44.29}/0.20 \\
\midrule
\multirow{3}{*}{iGSM} & ICL & 2.58/15.81 & 1.22/12.01 & 1.98/10.94 & 5.17/56.53 \\
 & Graph SFT & 45.19/46.40 & 44.83/46.50 & 43.26/43.97 & 55.43/55.12 \\
 & Graph-Based GRPO & \textbf{48.48}/\textbf{48.63} & \textbf{50.00}/\textbf{50.25} & \textbf{47.42}/\textbf{47.37} & \textbf{67.93}/\textbf{67.63} \\
\midrule
\multirow{3}{*}{MATH} & ICL & 2.56/9.74 & 6.07/7.58 & 4.92/\textbf{21.69} & 13.65/\textbf{89.06} \\
 & Graph SFT & 7.81/8.65 & 11.78/\textbf{15.29} & \textbf{12.17}/13.25 & 17.12/18.31 \\
 & Graph-Based GRPO & \textbf{12.14}/\textbf{12.44} & \textbf{13.52}/12.83 & 9.47/9.50 & \textbf{29.37}/29.82 \\
\midrule
\multirow{3}{*}{ProofWriter} & ICL & 56.00/65.45 & 25.20/4.75 & 28.50/46.10 & 82.75/92.35 \\
 & Graph SFT & 70.75/90.87 & 88.08/94.80 & 81.75/\textbf{94.90} & 86.88/94.01 \\
 & Graph-Based GRPO & \textbf{91.18}/\textbf{95.78} & \textbf{93.12}/\textbf{97.90} & \textbf{90.13}/93.43 & \textbf{98.91}/\textbf{97.76} \\
\midrule
\multicolumn{6}{l}{\textit{(b) $F_{\mathrm{ID}}$ / $F_{\mathrm{OOD}}$ / $F_{\mathrm{strong}}$ (\%).}} \\
\midrule
\multirow{3}{*}{GSM8K+} & ICL
& \textbf{0.29}/\textbf{0.36}/0.00
& \textbf{1.06}/0.18/0.00
& \textbf{1.61}/\textbf{0.21}/0.00
& \textbf{25.12}/0.16/0.00 \\
 & Graph SFT & 0.17/0.14/0.00 & 0.43/\textbf{0.36}/0.00 & 0.14/0.12/0.00 & 0.58/0.24/0.00 \\
 & Graph-Based GRPO & 0.14/0.17/0.00 & 0.24/0.21/0.00 & 0.28/0.17/0.00 & 0.55/\textbf{0.31}/0.00 \\
\midrule
\multirow{3}{*}{iGSM} & ICL & 4.67/56.38/1.83 & 6.08/55.06/3.17 & 5.15/58.21/2.38 & 6.38/64.56/1.09 \\
 & Graph SFT & 78.98/54.07/34.25 & 77.18/\textbf{67.48}/42.97 & 88.00/\textbf{67.28}/\textbf{51.11} & 86.61/60.27/47.77 \\
 & Graph-Based GRPO & \textbf{99.23}/\textbf{61.56}/\textbf{43.19} & \textbf{97.71}/65.68/\textbf{50.30} & \textbf{98.98}/60.97/50.81 & \textbf{90.44}/\textbf{64.69}/\textbf{48.32} \\
\midrule
\multirow{3}{*}{MATH} & ICL & 4.64/22.39/1.05 & 11.94/31.10/3.06 & 6.07/34.04/1.76 & 14.78/41.57/0.99 \\
 & Graph SFT & 34.31/39.06/15.71 & 35.21/44.85/18.31 & 43.84/44.17/21.16 & 54.31/52.07/30.45 \\
 & Graph-Based GRPO & \textbf{82.31}/\textbf{44.90}/\textbf{38.78} & \textbf{87.53}/\textbf{47.07}/\textbf{41.65} & \textbf{88.74}/\textbf{47.36}/\textbf{42.81} & \textbf{69.02}/\textbf{54.07}/\textbf{37.64} \\
\midrule
\multirow{3}{*}{ProofWriter} & ICL & 37.23/\textbf{88.05}/10.41 & 66.67/0.00/0.00 & 68.84/\textbf{87.63}/43.14 & 87.26/76.04/61.82 \\
 & Graph SFT & 74.57/65.94/\textbf{35.66} & 91.74/44.53/34.56 & 82.26/58.89/39.60 & 87.96/\textbf{100.00}/87.96 \\
 & Graph-Based GRPO & \textbf{93.93}/37.07/31.91 & \textbf{94.14}/\textbf{47.21}/\textbf{42.74} & \textbf{94.54}/49.59/\textbf{45.12} & \textbf{89.92}/\textbf{100.00}/\textbf{89.92} \\
\bottomrule
\end{tabular}
\captionof{table}{Per-model faithfulness results. \textbf{(a)} Internal vs.\ external
accuracy; a large gap indicates the emitted answer does not come from the
graph. \textbf{(b)} Intervention-based metrics. Values are percentages; bold
marks the best value of each metric within a model$\times$dataset across
methods. 
Entailment has no external answer; therefore,
neither internal--external accuracy nor
answer-based faithfulness metrics are applicable.
}
\label{tab:faithfulness_full}
\end{center}
\twocolumn

\clearpage
\section{Prompt for generating computational graph}\label{app:promt_for_graph}
\begin{figure}[H]
\centering

\begin{tcolorbox}[width=0.9\textwidth,colback=gray!5,colframe=black]

\begin{lstlisting}[basicstyle=\ttfamily\small,breaklines=true]
You are an expert in computational graphs and mathematical reasoning.

Given a math word problem, generate a computational graph as a JSON array where each node represents either:
1. A constant value (type: "const")
2. An arithmetic operation (type: "op")

Rules:
1. Each constant node has: {"id": <name>, "type": "const", "value": <number>}
2. Each operation node has: {"id": <result_name>, "type": "op", "op": <operation>, "inputs": [<input_ids>]}
3. Available operations: "add", "sub", "mul", "div", "pow", "mod"
4. Operation nodes are named after their result
5. List nodes in computational order (dependencies before usage)
6. Use descriptive names for node IDs
7. Constants should represent actual numbers from the problem
8. Each operation produces a new value
9. Each operation could only use previous const and op nodes as inputs
10. After the end of the graph, write "ANSWER: <number>"

EXAMPLE:
Question: Natalia sold clips to 48 of her friends in April, and then she sold half as many clips in May. How many clips did Natalia sell altogether in April and May?

[
  {"id": "april_clips", "type": "const", "value": 48},
  {"id": "half", "type": "const", "value": 0.5},
  {"id": "may_clips", "type": "op", "op": "mul", "inputs": ["april_clips", "half"]},
  {"id": "total_clips", "type": "op", "op": "add", "inputs": ["april_clips", "may_clips"]}
]
ANSWER: 72
\end{lstlisting}

\end{tcolorbox}

\caption{Example of prompt used for generating structured computational graphs with answer.}
\end{figure}
\label{}
\clearpage
\clearpage
\section{Task examples}
\label{app:graph_examples}

\begin{center}
\includegraphics[height=0.26\textheight]{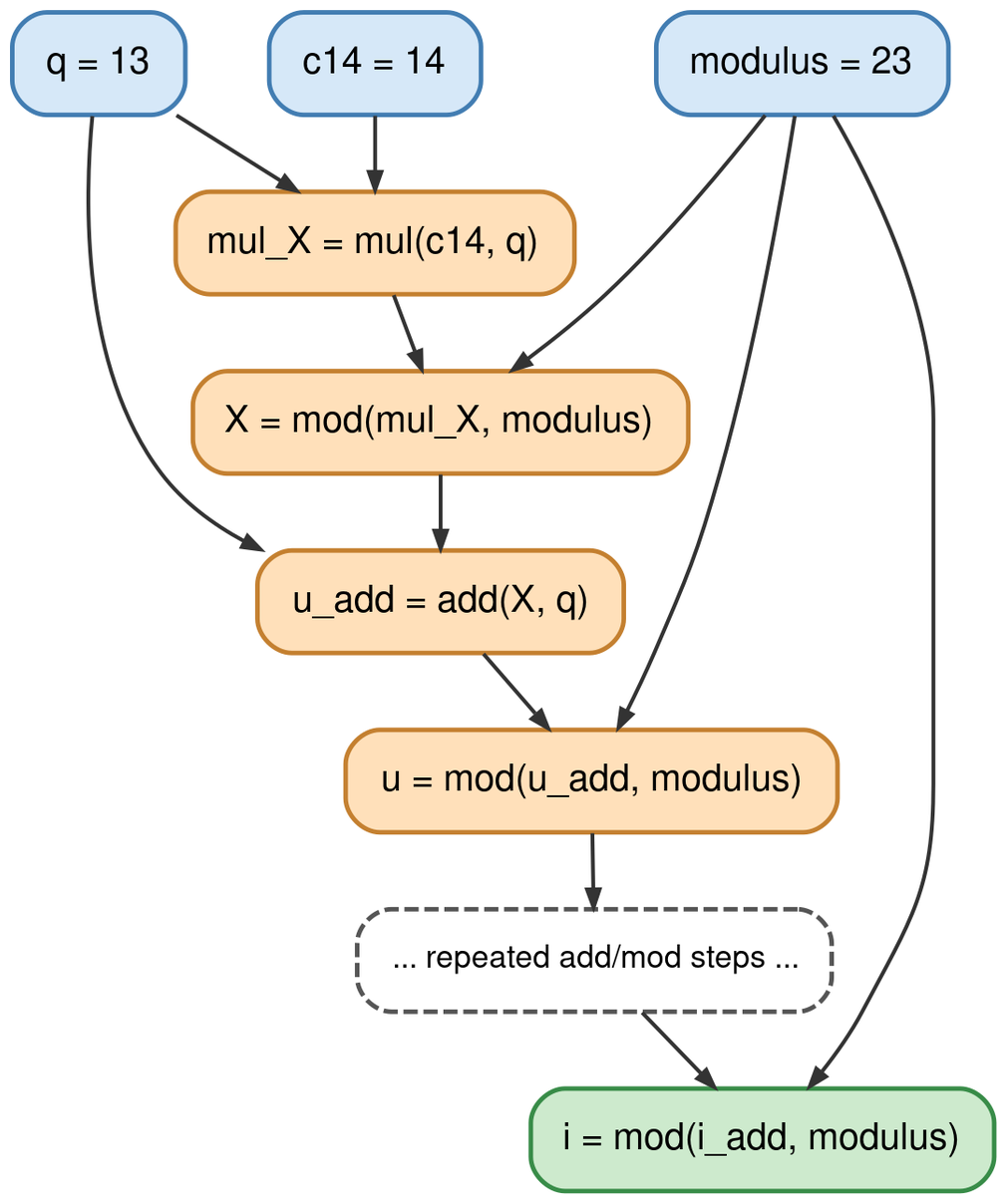}
\par\vspace{0.4em}
\parbox{0.85\textwidth}{\small\textbf{(a) iGSM.} (Shortened version of the original question) Seaside / Pine Valley /
Woodgrove lab counts linked by add/mod/mul relations under modular arithmetic.
How many Classrooms does Seaside Community College have? \textbf{Answer.} $10$.}

\vspace{1.6em}

\includegraphics[height=0.26\textheight]{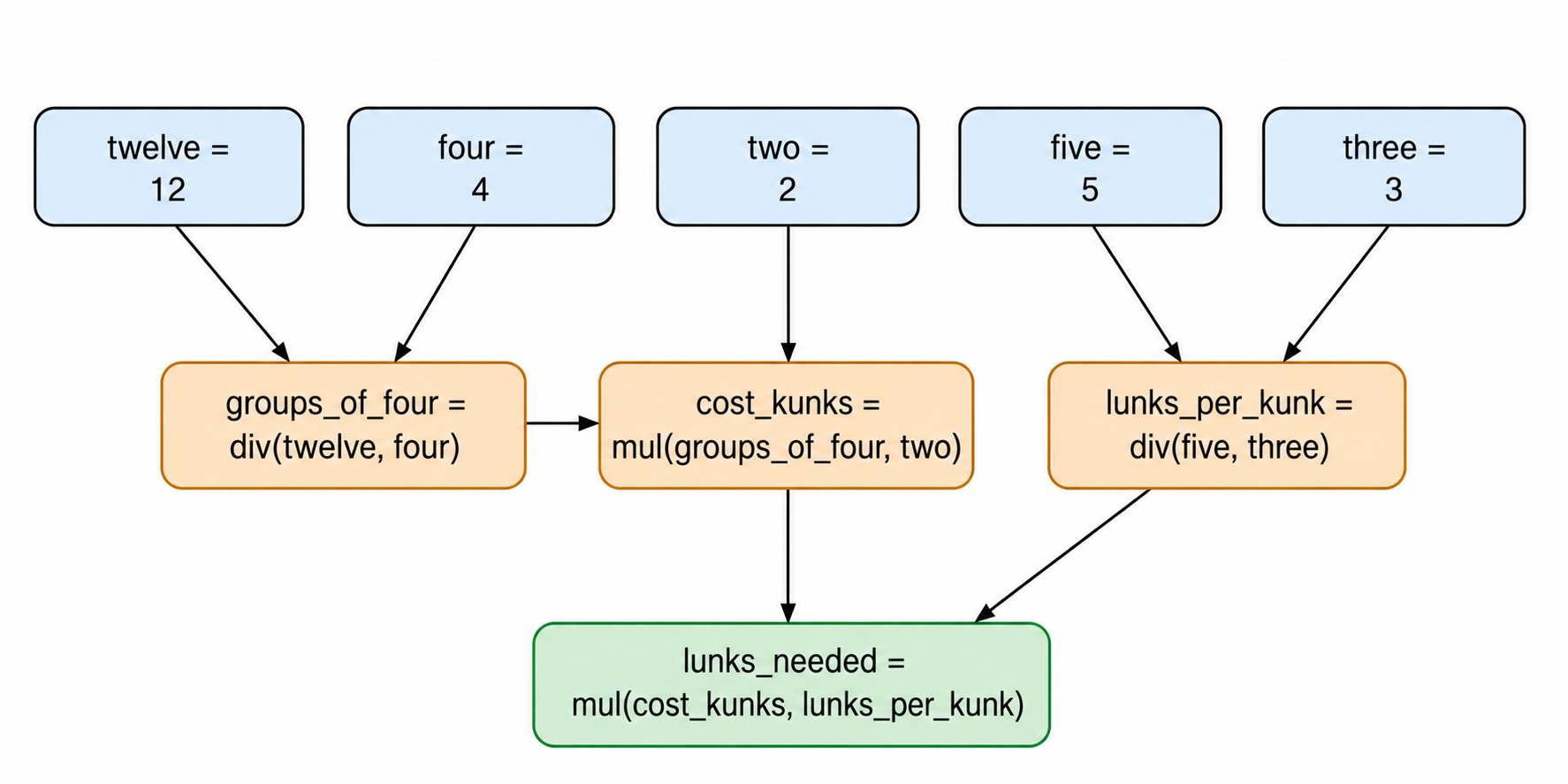}
\par\vspace{0.4em}
\parbox{0.85\textwidth}{\small\textbf{(b) MATH.} If $5$ lunks can be traded for $3$ kunks, and $2$ kunks will buy $4$ apples, how many lunks are needed to purchase one dozen apples?
\textbf{Answer.} $10$.}

\vspace{1.6em}

\includegraphics[height=0.26\textheight]{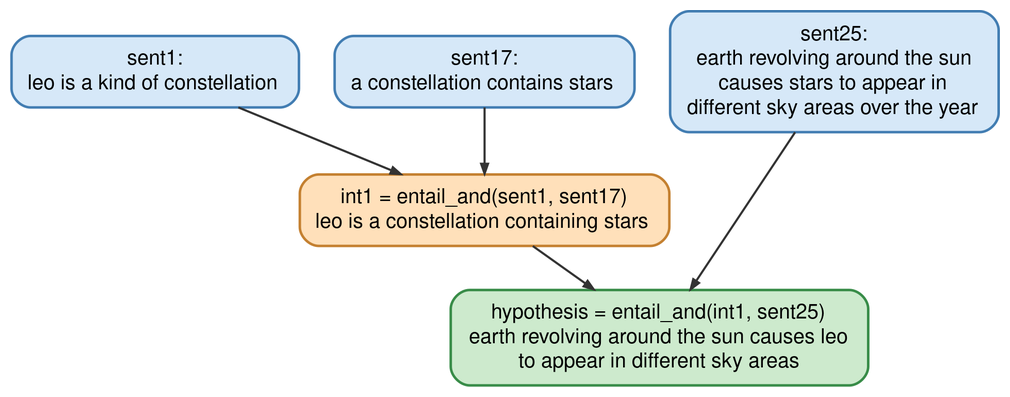}
\par\vspace{0.4em}
\parbox{0.85\textwidth}{\small\textbf{(c) Entailment.} Stars are organized into
patterns called constellations. One constellation is named Leo. Which statement
best explains why Leo appears in different areas of the sky throughout the year?
\textbf{Answer.} Earth revolves around the sun.}
\end{center}
\end{document}